\documentclass[11pt,a4paper]{article}
\usepackage[hyperref]{acl2020}
\usepackage{times}
\usepackage{latexsym}

\usepackage{microtype}

\aclfinalcopy 
\def\aclpaperid{367} 

\usepackage[T1]{fontenc}
\usepackage{dsfont}
\usepackage{colortbl}
\usepackage{xcolor}
\usepackage{enumitem}

\usepackage{mdframed}
\usepackage{booktabs}
\usepackage{graphicx}

\usepackage[normalem]{ulem}

\usepackage{xcolor}
\usepackage{soul}
\colorlet{soulred}{red!30}
\sethlcolor{soulred}%
\usepackage{amsmath,amsthm,amsfonts,amssymb,bm}

\usepackage{framed}
\usepackage{varioref}
\usepackage{float}
\usepackage{multirow}
\usepackage{dashrule}
\usepackage{url}

\title{Document-Level Event Role Filler Extraction \\ using Multi-Granularity Contextualized Encoding}

\author{Xinya Du \ {\normalfont and} \ Claire Cardie\\
  Department of Computer Science\\
  Cornell University \\
  Ithaca, NY, USA \\
  {\tt \{xdu, cardie\}@cs.cornell.edu} \\
  }

\date{}

\begin{document}
\maketitle
\begin{abstract}
Few works in the literature of event extraction have gone beyond individual sentences to make extraction decisions. This is problematic when the information needed to recognize an event argument is spread across multiple sentences.
We argue that document-level event extraction is a difficult task since it requires a view of a larger context to determine which spans of text correspond to event role fillers.
We first investigate how end-to-end neural sequence models (with pre-trained language model representations) perform on document-level role filler extraction, as well as how the length of context captured affects the models' performance.
To dynamically aggregate information captured by neural representations learned at different levels of granularity (e.g., the sentence- and paragraph-level), we propose a novel multi-granularity reader.
We evaluate our models on the MUC-4 event extraction dataset, and show that our best system performs substantially better than prior work. We also report findings on the relationship between context length and neural model performance on the task.
\end{abstract}

\section{Introduction}

\begin{figure}[t]
\centering
\small
\includegraphics[scale=0.5]{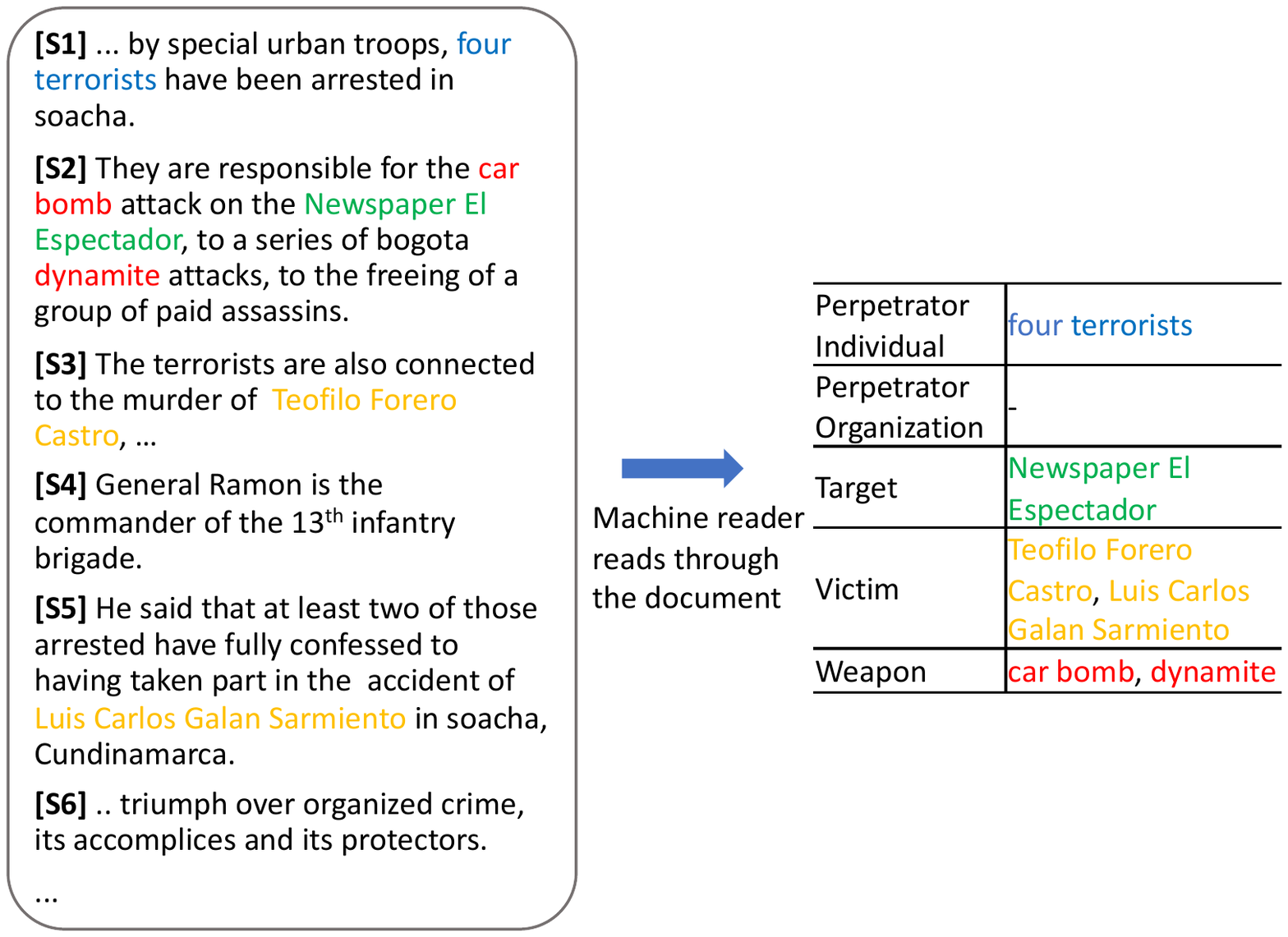}
\caption{The document-level event role fillers extraction task.}
\label{fig:e_task}
\end{figure}

The goal of document-level event extraction\footnote{The task is also referred to as \textit{template filling}~\cite{muc-1992-message}.} is to identify in an article events of a pre-specified type along with their event-specific role fillers, i.e., arguments.
The complete document-level extraction problem generally requires \textit{role filler extraction}, \textit{noun phrase coreference resolution} and \textit{event tracking} (i.e., determine which extracted role fillers belong to which event). 
In this work, we focus only on document-level role filler extraction. Figure \ref{fig:e_task} provides a representative example of this task. Given an article consisting of multiple paragraphs/sentences, and a fixed set of event types (e.g., terrorist events) and associated roles (e.g., \textsc{Perpetrator Individual}, \textsc{Victim}, \textsc{Weapon}), we aim to identify those spans of text that denote the role fillers for each event described in the text.
This generally requires both sentence-level understanding and accurate interpretation of the context beyond the sentence. Examples include identifying ``Teofilo Forero Castro'' (mentioned in S3) as a victim of the car bomb attack event (mentioned in S2), determining there's no role filler in S4 (both of which rely mainly on sentence-level understanding, and identifying ``four terrorists'' in S1 as a perpetrator individual (which requires coreference resolution across sentence boundaries).
Generating the document-level extractions for events is essential in facilitating downstream applications such as information retrieval and article summarization \cite{yang-mitchell-2016-joint}, and for real-life applications such as trends analysis of world events \cite{sundheim-1992-overview}.

Recent work in document-level event role filler extraction has employed a pipeline architecture with separate classifiers for each type of role and for relevant context detection \cite{patwardhan-riloff-2009-unified, huang-riloff-2011-peeling}. However these methods:
(1) suffer from error propagation across different pipeline stages;
and (2) require heavy feature engineering (e.g., lexico-syntactic pattern features for candidate role filler extraction; lexical bridge and discourse bridge features for detecting event-relevant sentences at the document level).
Moreover, the features are manually designed for a particular domain, which requires linguistic intuition and domain expertise \cite{nguyen-grishman-2015-event}.

Neural end-to-end models have been shown to excel at sentence-level information extraction tasks, such as named entity recognition \cite{lample-etal-2016-neural, chiu-nichols-2016-named} and ACE-type within-sentence event extraction \cite{chen-etal-2015-event, nguyen-etal-2016-joint, wadden-etal-2019-entity}. However, to the best of our knowledge, no prior work has investigated the formulation of document-level event role filler extraction as an end-to-end neural sequence learning task. 
In contrast to extracting events and their role fillers from standalone sentences, document-level event extraction poses special challenges for neural sequence learning models.
First, capturing long-term dependencies in long sequences remains a fundamental challenge for recurrent neural networks \cite{pmlr-v80-trinh18a}. To model long sequences, most RNN-based approaches use backpropagation through time. But it's still difficult for the models to scale to very long sequences. We provide empirical evidence for this for event extraction in Section 4.3.
Second, although pretrained bi-directional transformer models such as BERT \cite{devlin-etal-2019-bert} better capture long-distance dependencies as compared to an RNN architecture, they still have a constraint on the maximum length of the sequence, which is below the length of many articles about events.
%

In the sections below, we study how to train and apply end-to-end neural models for event role filler extraction. We first formalize the problem as a sequence tagging task over the tokens in a set of contiguous sentences in the document. To address the aforementioned challenges for neural models applied to long sequences,
(1) we investigate the effect of context length (i.e., maximum input segment length) on model performance, and find the most appropriate length;
and (2) 
propose a multi-granularity reader that dynamically aggregates the information learned from the local context (e.g., sentence-level) and the broader context (e.g., paragraph-level).
A quantitative evaluation and qualitative analysis of our approach on the MUC-4 dataset \cite{muc-1992-message} both show that the multi-granularity reader achieves substantial improvements over the baseline models and prior work.

For replication purposes, our repository for the evaluation and preprocessing scripts will be available at \url{https://github.com/xinyadu/doc_event_role}.


\section{Related Work}
Event extraction has been mainly studied under two paradigms: detecting the event trigger and extracting the arguments from an individual sentence (e.g., the ACE task~\cite{doddington-etal-2004-automatic}\footnote{\small\url{https://catalog.ldc.upenn.edu/LDC2006T06}}, vs. at the document level (e.g., the MUC-4 template-filling task \cite{sundheim-1992-overview}).

\paragraph{Sentence-level Event Extraction}
The ACE event extraction task requires extraction of the event trigger and its arguments from a sentence. For example, in the sentence `` ... Iraqi soldiers were \textit{killed} by U.S. artillery ...'', the goal is to identify the ``die'' event triggered by killed and the corresponding arguments (\textsc{place}, \textsc{victim}, \textsc{instrument}, etc.).
Many approaches have been proposed to improve performance on this specific task. \newcite{li-etal-2013-joint, li-etal-2015-improving-event} explore various hand-designed features; \newcite{nguyen-grishman-2015-event, nguyen-etal-2016-joint, chen-etal-2015-event, liu-etal-2017-exploiting, liu-etal-2018-jointly} employ deep learning based models such as recurrent neural networks (RNNs) and convolutional neural network (CNN). \newcite{wadden-etal-2019-entity} utilize pre-trained contextualized representations.
The approaches generally focus on sentence-level context for extracting event triggers and arguments and rarely generalize to the document-event extraction setting (Figure \ref{fig:e_task}).

Only a few models have gone beyond individual sentences to make decisions.
\newcite{ji-grishman-2008-refining} enforce event role consistency across documents.
\newcite{liao-grishman-2010-using} explore event type co-occurrence patterns to propagate event classification decisions. 
Similarly, \newcite{yang-mitchell-2016-joint} propose \textit{jointly} extracting events and entities within a document context.
Also related to our work are \newcite{duan-etal-2017-exploiting} and \newcite{zhao-etal-2018-document}, which utilize document embeddings to aid event detection with recurrent neural networks. Although these approaches make decisions with cross-sentence information, their extractions are still at the sentence level. 

\paragraph{Document-level Event Extraction}
has been studied mainly under the classic MUC paradigm \cite{muc-1992-message}. The full task involves the construction of answer key templates, one template per event (some documents in the dataset describe more than one events). Typically three steps are involved --- role filler extraction, role filler mention coreference resolution and event tracking). In this work we focus on role filler extraction.

From the modeling perspective, recent work explores both the local and additional context to make the role filler extraction decisions.
GLACIER \cite{patwardhan-riloff-2009-unified} jointly considers cross-sentence and noun phrase evidence in a probabilistic framework to extract role fillers.
TIER \cite{huang-riloff-2011-peeling} proposes to first determine the document genre with a classifier and then identify event-relevant sentences and role fillers in the document.
\newcite{huang2012modeling} propose a bottom-up approach that first aggressively identifies \textit{candidate} role fillers (with lexico-syntactic pattern features), and then removes the candidates that are in spurious sentences (i.e., not event-related) via a cohesion classifier (with discourse features).
Similar to \newcite{huang2012modeling}, we also incorporate both intra-sentence and cross-sentence features (paragraph-level features), but instead of using manually designed linguistic information, our models learn in an automatic way how to dynamically incorporate learned representations of the article. Also, in contrast to prior work that is pipeline-based, our approach tackles the task as an end-to-end sequence tagging problem.

There has also been work on unsupervised event schema induction \cite{chambers-jurafsky-2011-template, chambers-2013-event} and open-domain event extraction \cite{liu-etal-2019-open} from documents: the main idea is to group entities corresponding to the same role into an event template. Our models, on the other hand, are trained in supervised way and the event schemas are pre-defined.

Apart from event extraction, there has been increasing interest on cross-sentence relation extraction~\cite{mintz-etal-2009-distant, peng-etal-2017-cross, jia-etal-2019-document}. This work assumes that mentions are provided, and thus is more of a mention/entity-level classification problem. 
Our work instead focuses on role filler/span extraction using sequence tagging approaches; role filler type is determined during this process.

\paragraph{Capturing Long-term Dependencies for Neural Sequence Models}

For training neural sequence models such as RNNs, capturing long-term dependencies in sequences remains a fundamental challenge \cite{pmlr-v80-trinh18a}. Most approaches use backpropagation through time (BPTT) but it is difficult to scale to very long sequences. Many variations of models have been proposed to mitigate the effect of long sequence length, such as Long Short Term Memory (LSTM) Networks \cite{hochreiter1997long, gers1999learning, graves2013generating} and Gated Recurrent Unit Networks \cite{cho-etal-2014-learning}. Transformer based models \cite{vaswani2017attention, devlin-etal-2019-bert} have also shown improvements in modeling long text.
In our work for document-level event role filler extraction, we also implement LSTM layers in the models as well as utilize the pre-trained representations provided by the bi-directional transformer model -- BERT.
From an application perspective, we investigate the suitable length of context to incorporate for the neural sequence tagging model in the document-level extraction setting. We also study how to mitigate problems associated with long sequences by dynamically incorporating both sentence-level and paragraph-level representations in the model (Figure \ref{fig:multi}).


\begin{figure*}[t]
\centering
\small
\includegraphics[scale=0.65]{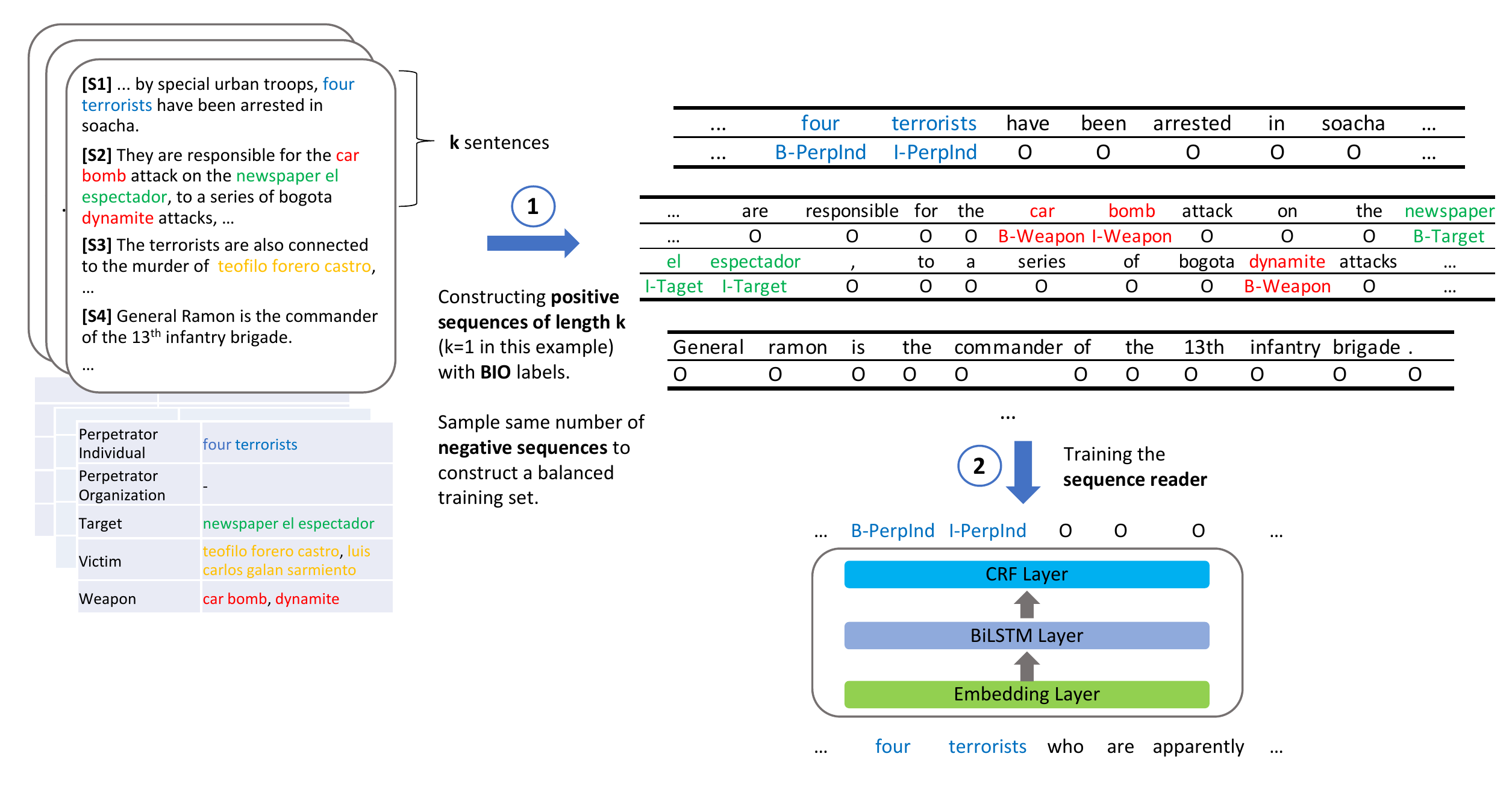}
\caption{An overview of our framework for training the sequence reader for event role filler extraction.}
\label{fig:general}
\end{figure*}

\section{Methodology}
In the following we describe (1) how we transform the document into paired token-tag sequences and formalize the task as a sequence tagging problem (Section 3.1); (2) the architectures of our base $k$-sentence reader (Section 3.2) and multi-granularity reader (Section 3.3).

\subsection{\small Constructing Paired Token-tag Sequences \\ \ \quad \quad from Documents and Gold Role Fillers}

We formalize document-level event role filler extraction as an end-to-end sequence tagging problem. The Figure \ref{fig:general} illustrates the general idea. Given a document and the text spans associated with the gold-standard (i.e., correct) fillers for each role, we adopt the BIO (Beginning, Inside, Outside) tagging scheme to transform the document into paired token/BIO-tag sequences..

We construct example sequences of variant context lengths for training and testing our end-to-end $k$-sentence readers (i.e., the single-sentence, double-sentence, paragraph and chunk readers).
By ``chunk'', we mean the chunk of contiguous sentences which is right within the sequence length constraint for BERT -- 512 in this case. 
Specifically, 
we use a sentence splitter\footnote{\url{https://spacy.io/}} to divide the document into sentences $s_1$, $s_2$, ..., $s_n$.
To construct the training set, starting from \textit{each} sentence $i$, we concatenate the $k$ contiguous sentences ($s_i$ to $s_{i+k-1}$) to form overlapping candidate sequences of length $k$ -- sequence 1 consists of $\{s_1, ..., s_k\}$, sequence 2 consists of $\{s_{2}, ..., s_{k+1}\}$, etc.
To make the training set balanced, we sample the same number of positive and negative sequences from the candidate sequences, where "positive" sequence contains at least one event role filler, and ``negative'' sequences contain no event role fillers.
To construct the dev/test set, where the reader is applied, we simply group the contiguous $k$ sentences together in order, producing $\frac{n}{k}$ sequences (i.e., sequence 1 consists of $\{s_1, ..., s_k\}$, sequence 2 consists of $\{s_{k+1}, ..., s_{2k}\}$, etc.)
For the paragraph reader, we set $k$ to average paragraph length for the training set, and to the real paragraph length for test set.

We denote the token in the sequence with $x$, the input for the $k$-sentence reader is $\mathbf{X}$ = $\{x^{(1)}_1, x^{(1)}_2, ..., x^{(1)}_{l_1}, ..., x^{(k)}_1, x^{(k)}_2, ..., x^{(k)}_{l_k}\}$; where $x^{(k)}_{i}$ is the $i$-th token of the $k$-th sentence, and $l_k$ is the length of the $k$-th sentence.

\subsection{$k$-sentence Reader}
\label{sec:kreader}
Since our general $k$-sentence reader does not recognize sentence boundaries, we simplify the notation for the input sequence as $\{x_1, x_2, ..., x_m\}$ here.

\paragraph{Embedding Layer}

In the embedding layer, we represent each token $x_i$ in the input sequence as the concatenation of its word embedding and contextual token representation:
\begin{itemize}
    \item \textit{Word Embedding}: We use the 100-dimensional GloVe pre-trained word embeddings \cite{pennington-etal-2014-glove} trained from 6B Web crawl data. We keep the pre-trained word embeddings fixed. Given a token $x_i$, we have its word embedding: $\mathbf{xe}_i = \mathbf{E}(x_i)$.
    
    \item \textit{Pre-trained LM representation}: Contextualized embeddings produced by pre-trained language models \cite{peters-etal-2018-deep, devlin-etal-2019-bert} have been proved to be capable of modeling context beyond the sentence boundary and improve performance on a variety of tasks. Here we employ the contextualized representations produced by \verb|BERT-base| for our $k$-sentence labeling model, as well as the multi-granularity reader to be introduced next.
    Specifically, we use the average of all the 12 layers' representations and freeze the weights \cite{peters-etal-2019-tune} during training after empirical trials\footnote{Using the representations of the last layer, or summing all the 12 layers' representations give consistently worse results.}. Given the sequence $\{x_1, x_2, ..., x_m\}$, we have:
    \begin{equation}
    \nonumber
    \mathbf{xb}_1, \mathbf{xb}_2, ..., \mathbf{xb}_m = \verb|BERT| (x_1, x_2, ..., x_m)
    \end{equation}
\end{itemize}
We forward the concatenation of the two representations for each token to the upper layers:
\begin{equation}
\nonumber
    \mathbf{x}_i = \verb|concat|(\mathbf{xe}_i, \mathbf{xb}_i)
\end{equation}

\paragraph{BiLSTM Layer}

To help the model better capture task-specific features between the sequence tokens. We use a multi-layer (3 layers) bi-directional LSTM encoder on top of the token representations, which we denote as $\verb|BiLSTM|$:
\begin{equation}
\begin{gathered}
\nonumber
    \{\mathbf{p}_1, \mathbf{p}_2,..., \mathbf{p}_m\} \\
    = \verb|BiLSTM|(\{\mathbf{x}_1, \mathbf{x}_2, ..., \mathbf{x}_m\})
\end{gathered}
\end{equation}

\paragraph{CRF Layer}
Drawing inspirations for sentence-level sequence tagging models on tasks like NER \cite{lample-etal-2016-neural}. Modeling the labeling decisions jointly rather than independently improves the models performance (e.g., the tag ``I-Weapon'' should not follow ``B-Victim''). We model labeling decisions jointly using a conditional random field \cite{lafferty2001conditional}.

After passing $\{\mathbf{p}_1, \mathbf{p}_2,..., \mathbf{p}_m\}$ through a linear layer, we have $\mathbf{P}$ of size $m \times$ size of tag space, where $\mathbf{P}_{i,j}$ is the score of the tag $j$ of the $i$-th token in the sequence. For a tag sequence $\mathbf{y} = \{y_1, ..., y_m\}$, we have the score for the sequence-tag pair as:

\begin{equation}
\nonumber
    score(\mathbf{X}, \mathbf{y}) = \sum_{i=0}^{m} \mathbf{A}_{y_i, y_{i+1}} + \sum_{i=1}^{m} \mathbf{P}_{i, y_i}
\end{equation}

$\mathbf{A}$ is the \textit{transition} matrix of scores such that $\mathbf{A}_{i,j}$ represents the score of a transition from the tag $i$ to tag $j$.
A \verb|softmax| function is applied over scores for all possible tag sequences, which yield a probability for the gold sequence $\mathbf{y}_{gold}$. The log-probability of the gold tag sequence is maximized during training. During decoding, the model predicts the output sequence that obtains the maximum score.

\begin{figure*}[!t]
\centering \small
\includegraphics[scale=0.7]{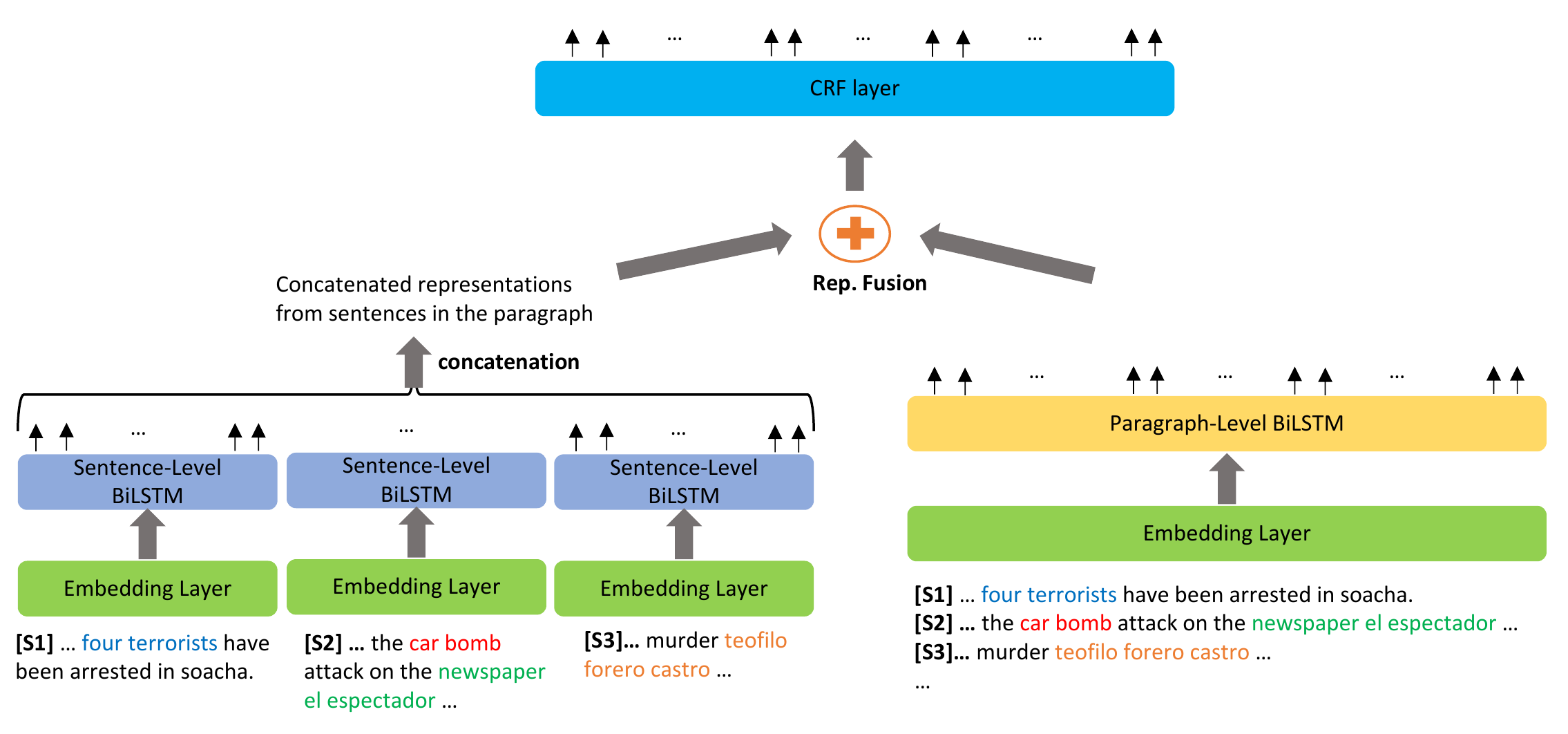}
\caption{Overview for our multi-granularity reader. The dark blue $\texttt{BiLSTM}_{sent.}$ produces sentence-level representations for each token, the yellow $\texttt{BiLSTM}_{para.}$ produces paragraph-level representations for each token.}
\label{fig:multi}
\end{figure*}

\subsection{Multi-Granularity Reader}
To explore the effect of aggregating contextualized token representations from different granularities (sentence- and paragraph-level), we propose the multi-granularity reader (Figure \ref{fig:multi}).

Similar to the general $k$-sentence reader, we use the same embedding layer here to represent the tokens. But we apply the embedding layer to two granularities of the paragraph text (sentence- and paragraph-level). Although the word embeddings are the same for the embedding layers from different granularities, the contextualized representations are different for each token -- when the token is encoded in the context of a sentence, or in the context of a paragraph.

Correspondingly, we build two BiLSTMs ($\verb|BiLSTM|_{sent.}$ and $\verb|BiLSTM|_{para.}$) on top of the sentence-level contextualized token representations $\{\mathbf{\tilde{x}}^{(1)}_1, ..., \mathbf{\tilde{x}}^{(1)}_{l_1}, ..., \mathbf{\tilde{x}}^{(k)}_{l_k}, ..., \mathbf{\tilde{x}}^{(k)}_{l_k}\}$, 
and the paragraph-level contextualized token representations $\{\mathbf{\hat{x}}^{(1)}_1, ..., \mathbf{\hat{x}}^{(1)}_{l_1}, ..., \mathbf{\hat{x}}^{(k)}_{l_k}, ..., \mathbf{\hat{x}}^{(k)}_{l_k}\}$:

\paragraph{Sentence-Level BiLSTM}
The $\verb|BiLSTM|_{sent.}$ is applied sequentially to each sentence in the paragraph:
\begin{equation}
\begin{gathered}
\nonumber
\{\mathbf{\tilde{p}}^{(1)}_1, \mathbf{\tilde{p}}^{(1)}_2,..., \mathbf{\tilde{p}}^{(1)}_{l_1}\} \\
= \verb|BiLSTM|_{sent.}(\{\mathbf{\tilde{x}}^{(1)}_1, \mathbf{\tilde{x}}^{(1)}_2, ..., \mathbf{\tilde{x}}^{(1)}_{l_1}\}) \\
... \\
\{\mathbf{\tilde{p}}^{(k)}_1, \mathbf{\tilde{p}}^{(k)}_2, ..., \mathbf{\tilde{p}}^{(k)}_{l_k}\} 
\\ = \verb|BiLSTM|_{sent.}(\{\mathbf{\tilde{x}}^{(k)}_1, \mathbf{\tilde{x}}^{(k)}_2, ..., \mathbf{\tilde{x}}^{(k)}_{l_k}\}) \\
\end{gathered}
\end{equation}

Then we have the sentence-level representations for each token in the paragraph as $\{\mathbf{\tilde{p}}^{(1)}_1,..., \mathbf{\tilde{p}}^{(1)}_{l_1}, ..., \mathbf{\tilde{p}}^{(k)}_1, ..., \mathbf{\tilde{p}}^{(k)}_{l_k}\}$

\paragraph{Paragraph-Level BiLSTM}
Another BiLSTM layer ($\verb|BiLSTM|_{para.}$) is applied to the \textit{entire} paragraph (as compared to $\verb|BiLSTM|_{sent.}$, which is applied to each sentence), to capture the dependency between tokens in the paragraph:

\begin{equation}
\begin{gathered}
\nonumber
\{\mathbf{\hat{p}}^{(1)}_1,..., \mathbf{\hat{p}}^{(1)}_{l_1}, ..., \mathbf{\hat{p}}^{(k)}_1, ..., \mathbf{\hat{p}}^{(k)}_{l_k}\} \\ 
= \verb|BiLSTM|_{para.}(\{\mathbf{
\hat{x}}^{(1)}_1, ..., \mathbf{\hat{x}}^{(1)}_{l_1}, ..., \mathbf{\hat{x}}^{(k)}_{l_k}, ..., \mathbf{\hat{x}}^{(k)}_{l_k}\})
\end{gathered}
\end{equation}

\paragraph{Fusion and Inference Layer}

For each token $x^{(j)}_i$ (the $i$-th token in the $j$-th sentence), to fuse the representations learned at the sentence-level ($\mathbf{\tilde{p}}^{(j)}_i$) and paragraph-level ($\mathbf{\hat{p}}^{(j)}_i$), we propose two options -- the first uses a sum operation, and the second uses a \textit{gated} fusion operation:

\begin{itemize}
    \item \textit{Simple Sum Fusion}: 
    \begin{equation}
    \label{equ:sum_fusion}
    \nonumber
        \mathbf{p}^{(j)}_i = \mathbf{\tilde{p}}^{(j)}_i + \mathbf{\hat{p}}^{(j)}_i
    \end{equation}
    
    \item \textit{Gated Fusion}: The gated fusion compute the gate vector $\mathbf{g}^{(j)}_i$ with its sentence-level token representation $\mathbf{\tilde{p}}^{(j)}_i$ and paragraph-level token representation $\mathbf{\hat{p}}^{(j)}_i$, to control how much information should be incorporated from the two representations.
    \begin{equation}
    \begin{gathered}
    \nonumber
    \label{equ:gate_fusion}
        \mathbf{g}^{(j)}_i = sigmoid(\mathbf{W}_1 \mathbf{\tilde{p}}^{(j)}_i + \mathbf{W}_2 \mathbf{\hat{p}}^{(j)}_i + b)\\
        \mathbf{p}^{(j)}_i = \mathbf{g}^{(j)}_i \odot \mathbf{\tilde{p}}^{(j)}_i + (1 - \mathbf{g}^{(j)}_i) \odot \mathbf{\hat{p}}^{(j)}_i \\
        \odot: \textnormal{element-wise product}
    \end{gathered}
    \end{equation}
\end{itemize}

Similarly to in the general $k$-sentence reader, we add the CRF layer (section \ref{sec:kreader}) on top of the fused representations for each token in the paragraph $\{\mathbf{p}^{(1)}_1,..., \mathbf{p}^{(1)}_{l_1}, ..., \mathbf{p}^{(k)}_1, ..., \mathbf{p}^{(k)}_{l_k}\}$, to help jointly model the labeling decisions between tokens in the paragraph.

\section{Experiments and Analysis}

We evaluate our models' performance on the MUC-4 event extraction benchmark \cite{muc-1992-message}, and compare to prior work. We also report findings on the effect of context length on the end-to-end readers' performance on this document-level task.

\subsection{Dataset and Evaluation Metrics}

\paragraph{MUC-4 Event Extraction Dataset} 
The MUC-4 dataset consists of 1,700 documents with associated answer key (role filler) templates. To make sure our results are comparable to the previously reported results on this dataset, we use the 1300 documents for training, 200 documents (\verb|TST1+TST2|) as the development set and the 200 documents (\verb|TST3+TST4|) as the test set.

\paragraph{Evaluation Metrics}
Following the prior work, we use \textit{head noun phrase match} to compare the extractions against gold role fillers for evaluation
\footnote{Duplicate role fillers (i.e., extractions for the same role that have the same head noun) are conflated before being scored; they are counted as one hit (if the system produces it) or one miss (if the system fails to produce any of the duplicate mentions).};
besides noun phrase matching, we also report \textit{exact match} accuracy, to capture how well the models are capturing the role fillers' boundary\footnote{Similarly, duplicate extractions with the same string are counted as one hit or miss.}.
Our results are reported as Precision (P), Recall (R) and F-measure (F-1) score for the macro average for all the event roles. In Table \ref{tab:per}, we also present the scores for each event role (i.e., \textsc{perpetrator individuals}, \textsc{perpetrator organizations}, \textsc{physical targets}, \textsc{victims} and \textsc{weapons}) based on the head noun match metric.
The detailed documentation and implementation for the evaluation script will be released.

\subsection{Baseline Systems and Our Systems}

\begin{table*}[!t]
\small
\centering
\begin{tabular}{l|ccc|ccc}
\toprule
 & \multicolumn{3}{c|}{Head Noun Match} & \multicolumn{3}{c}{Exact Match} \\ \cmidrule(lr){2-7}
& Prec.    & Recall    & F-1      & Prec.   & Recall   & F-1    \\ \midrule
GLACIER \cite{patwardhan-riloff-2009-unified}  & 47.80 & 57.20 & 52.08 & -     & -     & -     \\   
TIER \cite{huang-riloff-2011-peeling}   & 50.80 & 61.40 & 55.60 & -     & -     & -     \\
Cohesion Extract \cite{huang2012modeling} & 57.80 & 59.40 & 58.59 & -     & -     & -  \\ \midrule
 \multicolumn{1}{c|}{\textit{w/o contextualized embedding}} & \\    
Single-Sentence Reader  & 48.69 & 56.11 & 52.14 & 46.16 & 53.16 & 49.41 \\
Double-sentence Reader  & 56.37 & 47.53 & 51.57 & 53.70 & 43.95 & 48.34 \\
Paragraph Reader  & 53.19 & 53.16 & 53.17 & 49.45 & 49.26 & 49.35 \\
Chunk Reader      & \cellcolor{blue!25}\textbf{61.76} & 37.04 & 46.31 & \cellcolor{blue!25}\textbf{56.91} & 34.92 & 43.28 \\ \midrule
 \multicolumn{1}{c|}{\textit{w/ contextualized embedding}} & \\    
Contextualized Single-Sentence Reader   & 47.32 & 61.26 & 53.39 & 44.40 & \cellcolor{blue!25}\textbf{57.67} & 50.17 \\
Contextualized Double-sentence Reader  & 57.17 & 53.36 & 55.20 & 53.38 & 49.22 & 51.22 \\
Contextualized Paragraph Reader  & 56.78 & 52.64 & 54.64 & 53.36 & 49.65 & 51.44 \\
Contextualized Chunk Reader  & 60.90 & 41.10 & 49.07 & 55.18 & 37.51 & 44.66 \\ \midrule
Multi-Granularity Reader & 56.44 & \cellcolor{blue!25}\textbf{62.77} & \cellcolor{blue!25}\textbf{59.44} & 52.03 & 56.81 & \cellcolor{blue!25}\textbf{54.32} \\
\bottomrule
\end{tabular}
\caption{Macro average results for the document-level event extraction task {(highest number of the column boldfaced).}}
\label{tab:macro}
\end{table*}

We compare to the pipeline and manual feature engineering based systems:
\textbf{GLACIER} \cite{patwardhan-riloff-2009-unified} consists of a sentential event classifier and a set of plausible role filler recognizers for each event role. The final extraction decisions are based on the product of normalized sentential and phrasal probabilities;
\textbf{TIER} \cite{huang-riloff-2011-peeling} proposes a multi-stage approach. It processes a document in three stages: classifying narrative document, recognizing event sentence and noun phrase analysis.
\textbf{Cohesion Extract} \cite{huang2012modeling} adopts a bottom-up approach, which first aggressively identifies candidate role fillers in the document and then refines the candidate set with cohesion sentence classifier. Cohesion Extract obtains substantially better precision and with similar level of recall as compared to GLACIER and TIER.

To investigate how the neural models capture the long dependency in the context of variant length (single-sentence, double-sentence, paragraph or longer), we initialize the $k$ in $k$-sentence reader to different values to build the:
\textbf{Single-Sentence Reader} ($k=1$), which reads through the document sentence-by-sentence to extract the event role fillers;
\textbf{Double-Sentence Reader} ($k=2$), which reads the document with step of two sentences;
\textbf{Paragraph Reader} ($k= \#$ sentences in the paragraph), which reads the document paragraph-by-paragraph;
\textbf{Chunk Reader} ($k=$ maximum $\#$ of sentences that fit right in the length constraint for pretrained LM models), which reads the document with the longest step (the constraint of BERT model).

The final row in Table \ref{tab:macro}\&\ref{tab:per} presents the results obtained with our \textbf{Multi-Granularity Reader}. Similar to the paragraph-level reader, it reads through document paragraph-by-paragraph, but learns the representations for both intra-sentence and inter-sentence context.

\subsection{Results and Findings}

We report the macro average results in Table \ref{tab:macro}. To understand in detail how the models extract the fillers for \textit{each} event role, we also report the per event role results in Table \ref{tab:per}. We summarize the results into important findings below:

\begin{table*}[!t]
\centering
\resizebox{\textwidth}{!}{
\begin{tabular}{l|ccc|ccc|ccc|ccc|ccc}
\toprule 
& \multicolumn{3}{c|}{PerpInd} & \multicolumn{3}{c|}{PerpOrg} & \multicolumn{3}{c|}{Target} & \multicolumn{3}{c|}{Victim} & \multicolumn{3}{c}{Weapon}  \\ \cmidrule(lr){2-16}
 & P       & R       & F-1     & P       & R       & F-1     & P       & R       & F-1    & P       & R       & F-1    & P       & R       & F-1       \\ \midrule
\begin{tabular}[c]{@{}l@{}}GLACIER \\ \cite{patwardhan-riloff-2009-unified} \end{tabular}        & 51    & 58    & 54    & 34    & 45    & 38    & 42    & 72    & 53    & 55    & 58    & 56    & 57    & 53    & 55         \\
\begin{tabular}[c]{@{}l@{}}TIER \\  \cite{huang-riloff-2011-peeling}  \end{tabular}
& 54    & 57    & 56    & 55    & 49    & 51    & 55    & 68    & 61    & 63    & 59    & 61    & 62    & 64    & 63           \\
\begin{tabular}[c]{@{}l@{}}Cohesion Extract \\   \cite{huang2012modeling}   \end{tabular}
& 54      & 57      & \cellcolor{blue!25}\textbf{56}      & 55      & 49      & 51      & 55      & 68      & 61     & 63      & 59      & \cellcolor{blue!25}\textbf{61}     & 62      & 64      & 63    \\ \midrule
 \multicolumn{1}{c|}{ \emph{w/o contextualized embedding}} & \\                     
Single-Sentence Reader   &38.38 & 50.68 & 43.68 & 40.98 & 69.05 & 51.44 & 62.50 & 42.76 & 50.78 & 36.69 & 55.79 & 44.27 & 64.91 & 62.30 & 63.58 \\
Double-Sentence Reader   & 50.00 & 35.14 & 41.27 & 63.83 & 35.71 & 45.80 & 61.62 & 44.83 & 51.90 & 51.02 & 54.74 & 52.81 & 55.41 & 67.21 & 60.74 \\
Paragraph Reader & 42.51 & 51.35 & 46.52 & 44.80 & 54.76 & 49.28 & 70.33 & 43.45 & 53.71 & 53.75 & 47.37 & 50.36 & 54.55 & 68.85 & 60.87 \\
Chunk Reader  & 65.63 & 26.19 & 37.44 & 50.00 & 45.45 & 47.62 & 77.78 & 22.62 & 35.05 & 55.00 & 21.15 & 30.56 & 60.42 & 69.77 & 64.76 \\ \midrule
 \multicolumn{1}{c|}{\emph{w/ contextualized embedding}} & \\ 
C-Single-Sentence Reader  & 44.97 & 52.70 & 48.53 & 35.15 & 73.81 & 47.62 & 71.74 & 24.83 & 36.89 & 33.63 & 77.89 & 46.98 & 51.11 & 77.05 & 61.46 \\
C-Double-Sentence Reader   & 63.49 & 31.76 & 42.34 & 53.25 & 48.81 & 50.93 & 69.52 & 50.34 & 58.40 & 44.03 & 62.11 & 51.53 & 55.56 & 73.77 & 63.38 \\
C-Paragraph Reader & 43.92 & 53.38 & 48.19 & 52.94 & 54.76 & 53.84 & 74.19 & 44.83 & 55.89 & 50.57 & 46.32 & 48.35 & 62.30 & 63.93 & 63.10 \\
C-Chunk Reader & 57.14 & 27.38 & 37.02 & 47.62 & 40.91 & 44.01 & 70.27 & 29.76 & 41.81 & 59.46 & 42.31 & 49.44 & 70.00 & 65.12 & 67.47 \\ \midrule
Multi-Granularity Reader  & 53.08 & 52.23 & 52.65 & 50.99 & 67.88 & \cellcolor{blue!25}\textbf{58.23} & 60.38 & 64.10 & \cellcolor{blue!25}\textbf{62.18} & 49.34 & 62.05 & 54.97 & 68.42 & 67.57 & \cellcolor{blue!25}\textbf{67.99} \\
\bottomrule
\end{tabular}}
\caption{Per event role results based on head noun match metric ({\small``C-'' stands for contextualized}). { The highest F-1 are boldfaced for each event role.}}
\label{tab:per}
\end{table*}

\begin{table*}[ht]
\small
\centering
\begin{tabular}{l|ccc|ccc}
\toprule
& \multicolumn{3}{c|}{Head Noun Match} & \multicolumn{3}{|c}{Exact Match} \\ \cmidrule(lr){2-7}
& Precision    & Recall    & F-1      & Precision   & Recall   & F-1    \\ \midrule
Multi-granularity Reader & 56.44 & 62.77 & \textbf{59.44} & 52.03 & 56.81 & \textbf{54.32}  \\ \midrule
\ \ \ w/o gated fusion & 48.09 & 67.32 & 56.10 & 43.75 & 62.37 & 51.43 \\
\ \ \ w/o BERT & 59.16 & 50.80 & 54.66 & 55.48 & 46.99 & 50.88 \\
\ \ \ w/o CRF layer & 50.52 & 56.95 & 53.54 & 47.02 & 53.55 & 50.07 \\ 
\bottomrule
\end{tabular}
\caption{Ablation study on modules' influence on the multi-granularity reader.}
\label{tab:ablation}
\end{table*}

\begin{itemize}[]
    \item \textit{The end-to-end neural readers can achieve nearly the same level or significantly better results than the pipeline systems.}
    Although our models rely on no hand-designed features, the contextualized double-sentence reader and paragraph reader achieves nearly the same level of F-1 compared to Cohesion Extraction (CE), judging by the head noun matching metric. Our multi-granularity reader performs significantly better ($\sim$60) than the prior state-of-the-art.
    \item \textit{Contextualized embeddings for the sequence consistently improve the neural readers' performance.}
    The results show that the contextualized $k$-sentence readers all outperform their non-contextualized counterparts, especially when $k>1$. The trends also exhibit in the per event role analysis (Table \ref{tab:per}).
    To notice, we freeze the transformers' parameters during training ({fine-tuning yields worse results}).
    \item \textit{It's not the case that modeling the longer context will result in better neural sequence tagging model on this document-level task.} 
    When increasing the input context from a single sentence to two sentences, the reader has a better precision and lower recall, resulting in no better F-1; 
    When increase the input context length further to the entire paragraph, the precision increases and recall remains the same level, resulting in higher F-1;
    When we keep increasing the length of input context, the reader becomes more conservative and F-1 drops significantly.
    All these indicate that focusing on the local (intra-sentence) and broader (paragraph-level) context are both important for the task.
    Similar results regarding the context length have also been found in document-level coreference resolution~\cite{joshi2019bert}.
    \item \textit{Our multi-granularity reader that dynamically incorporates sentence-level and paragraph-level contextual information performs significantly better}, than the non-end-to-end systems and our base $k$-sentence readers on the macro average F-1 metric. In terms of the per event role performance (Table \ref{tab:per}), our reader: (1) substantially outperforms CE with a $\sim7$ F-1 gap on the \textsc{Perpetrator Organization} role; (2) slightly outperforms CE ($\sim$1 on the Target category); (3) achieves nearly the same-level of F-1 for \textsc{Perpetrator Individual} and worse F-1 on \textsc{Victim} category.
\end{itemize}

\section{Further Analysis}

We conduct an ablation study on how modules of our multi-granularity reader affect its performance on this document-level extraction task (Table \ref{tab:ablation}). 
From the results, we find that:
(1) when replacing the gated fusion operation with the simple sum of the sentence- and paragraph-level token representations, the precision and F-1 drop substantially, which proves the importance of dynamically incorporating context;
(2) when removing the BERT's contextualized representations, the model becomes more conservative and yields substantially lower recall and F-1;
(3) when replacing the CRF layer and make independent labeling decisions for each token, both the precision and recall drops substantially.

We also do an error analysis with examples and predictions from different models, to understand qualitatively the advantages and disadvantages of our models. 
In the first example below ({\small \colorlet{soulred}{green!50}\hl{green span}: gold extraction, the \textcolor{green}{role} after is the span's event role}), the multi-granularity (MG) reader and single-sentence reader correctly extracts the two target expressions, which the paragraph reader overlooks. Although only in the last sentence the attack and targets are mentioned, our MG reader successfully captures this with focusing on both the paragraph-level and intra-sentence context.
\begin{quote}
\small
... the announcer says president virgilio barco will tonight disclose his government's peace proposal. ...... .  Near the end, the announcer adds to the initial report on the el tomate attack with a 3-minute update that adds 2 injured, \colorlet{soulred}{green!50}\hl{21 houses} \textcolor{green} {\small Target} destroyed, and \colorlet{soulred}{green!50}\hl{1 bus} \textcolor{green}{Target} burned.
\end{quote}

In the second example ({\small \colorlet{soulred}{red!50}\hl{red span}: false positive perpInd extraction by the single-sentence reader}), although ``members of the civil group'' appears in a sentence about explosion, judging from paragraph-level context or reasoning about the expression itself should help confirm that it is not perpetrator individual. The MG and paragraph reader correctly handles this and also extracts ``the bomb''.
\begin{quote}
\small
.... An attack came at approximately 22:30 last night. \colorlet{soulred}{red!50}\hl{Members of the civil group} and the peruvian investigative police went to the site of the explosion.
The members of the republican guard antiexplosives brigade are investigating to determine the magnitude of \colorlet{soulred}{green!50}\hl{the bomb} \textcolor{green}{Weapon} used in this attack. 
\end{quote}

There's substantial improvement space for our MG reader's predictions. There are many role fillers which the reader overlooks. In the example below, ``La Tandona'' being a perpetrator organization is implicitly expressed in the document and the phrase did not appear elsewhere in the corpus. But external knowledge (e.g., Wikipedia) could help confirm its event role.
\begin{quote}
\small ... Patriotic officer, it is time we sit down to talk, to see what we can do with our fatherland, and what are we going to do with \colorlet{soulred}{green!50}\hl{La Tandona} \textcolor{green}{PerpOrg}. ....  To continue defending what, we ask you. ... . 
\end{quote}

In the last example, there are no explicit expression such as ``kill'' or ``kidnap'' in the context for the target. Thus it requires deeper understanding of the \textit{entire} narrative and \textit{reasoning} about the surrounding context to understand that ``Jorge Serrano Gonzalez'' is involved in a terrorism event.
\begin{quote}
\small ...  said that the guerrillas are desperate and 
... .  The president expressed his satisfaction at the release of Santander department senator \colorlet{soulred}{green!50}\hl{Jorge Serrano Gonzalez} \textcolor{green}{Target}, whom he described as one of the most important people that colombian democracy has at this moment.
\end{quote}

\section{Conclusion and Future Work}
We have demonstrated that document-level event role filler extraction could be successfully tackled with end-to-end neural sequence models.
Investigations on how the input context length affects the neural sequence readers' performance show that context of very long length might be hard for the neural models to capture and results in lower performance.
We propose a novel multi-granularity reader to dynamically incorporate paragraph- and sentence-level contextualized representations. Evaluations on the benchmark dataset and qualitative analysis prove that our model achieves substantial improvement over prior work.
In the future work, it would be interesting to further explore how the model can be adapted to jointly extract role fillers, tackles coreferential mentions and constructing event templates.

\section*{Acknowledgments}
We thank the anonymous reviewers and Ana Smith for helpful feedback.

\bibliography{acl2020}

\begin{thebibliography}{39}
\expandafter\ifx\csname natexlab\endcsname\relax\def\natexlab#1{#1}\fi

\bibitem[{Chambers(2013)}]{chambers-2013-event}
Nathanael Chambers. 2013.
\newblock \href {https://www.aclweb.org/anthology/D13-1185} {Event schema
  induction with a probabilistic entity-driven model}.
\newblock In \emph{Proceedings of the 2013 Conference on Empirical Methods in
  Natural Language Processing}, pages 1797--1807, Seattle, Washington, USA.
  Association for Computational Linguistics.

\bibitem[{Chambers and Jurafsky(2011)}]{chambers-jurafsky-2011-template}
Nathanael Chambers and Dan Jurafsky. 2011.
\newblock \href {https://www.aclweb.org/anthology/P11-1098} {Template-based
  information extraction without the templates}.
\newblock In \emph{Proceedings of the 49th Annual Meeting of the Association
  for Computational Linguistics: Human Language Technologies}, pages 976--986,
  Portland, Oregon, USA. Association for Computational Linguistics.

\bibitem[{Chen et~al.(2015)Chen, Xu, Liu, Zeng, and
  Zhao}]{chen-etal-2015-event}
Yubo Chen, Liheng Xu, Kang Liu, Daojian Zeng, and Jun Zhao. 2015.
\newblock \href {https://doi.org/10.3115/v1/P15-1017} {Event extraction via
  dynamic multi-pooling convolutional neural networks}.
\newblock In \emph{Proceedings of the 53rd Annual Meeting of the Association
  for Computational Linguistics and the 7th International Joint Conference on
  Natural Language Processing (Volume 1: Long Papers)}, pages 167--176,
  Beijing, China. Association for Computational Linguistics.

\bibitem[{Chiu and Nichols(2016)}]{chiu-nichols-2016-named}
Jason~P.C. Chiu and Eric Nichols. 2016.
\newblock \href {https://doi.org/10.1162/tacl_a_00104} {Named entity
  recognition with bidirectional {LSTM}-{CNN}s}.
\newblock \emph{Transactions of the Association for Computational Linguistics},
  4:357--370.

\bibitem[{Cho et~al.(2014)Cho, van Merri{\"e}nboer, Gulcehre, Bahdanau,
  Bougares, Schwenk, and Bengio}]{cho-etal-2014-learning}
Kyunghyun Cho, Bart van Merri{\"e}nboer, Caglar Gulcehre, Dzmitry Bahdanau,
  Fethi Bougares, Holger Schwenk, and Yoshua Bengio. 2014.
\newblock \href {https://doi.org/10.3115/v1/D14-1179} {Learning phrase
  representations using {RNN} encoder{--}decoder for statistical machine
  translation}.
\newblock In \emph{Proceedings of the 2014 Conference on Empirical Methods in
  Natural Language Processing ({EMNLP})}, pages 1724--1734, Doha, Qatar.
  Association for Computational Linguistics.

\bibitem[{Devlin et~al.(2019)Devlin, Chang, Lee, and
  Toutanova}]{devlin-etal-2019-bert}
Jacob Devlin, Ming-Wei Chang, Kenton Lee, and Kristina Toutanova. 2019.
\newblock \href {https://doi.org/10.18653/v1/N19-1423} {{BERT}: Pre-training of
  deep bidirectional transformers for language understanding}.
\newblock In \emph{Proceedings of the 2019 Conference of the North {A}merican
  Chapter of the Association for Computational Linguistics: Human Language
  Technologies, Volume 1 (Long and Short Papers)}, pages 4171--4186,
  Minneapolis, Minnesota. Association for Computational Linguistics.

\bibitem[{Doddington et~al.(2004)Doddington, Mitchell, Przybocki, Ramshaw,
  Strassel, and Weischedel}]{doddington-etal-2004-automatic}
George Doddington, Alexis Mitchell, Mark Przybocki, Lance Ramshaw, Stephanie
  Strassel, and Ralph Weischedel. 2004.
\newblock \href {http://www.lrec-conf.org/proceedings/lrec2004/pdf/5.pdf} {The
  automatic content extraction ({ACE}) program {--} tasks, data, and
  evaluation}.
\newblock In \emph{Proceedings of the Fourth International Conference on
  Language Resources and Evaluation ({LREC}{'}04)}, Lisbon, Portugal. European
  Language Resources Association (ELRA).

\bibitem[{Duan et~al.(2017)Duan, He, and Zhao}]{duan-etal-2017-exploiting}
Shaoyang Duan, Ruifang He, and Wenli Zhao. 2017.
\newblock \href {https://www.aclweb.org/anthology/I17-1036} {Exploiting
  document level information to improve event detection via recurrent neural
  networks}.
\newblock In \emph{Proceedings of the Eighth International Joint Conference on
  Natural Language Processing (Volume 1: Long Papers)}, pages 352--361, Taipei,
  Taiwan. Asian Federation of Natural Language Processing.

\bibitem[{Gers et~al.(1999)Gers, Schmidhuber, and Cummins}]{gers1999learning}
Felix~A Gers, J{\"u}rgen Schmidhuber, and Fred Cummins. 1999.
\newblock Learning to forget: Continual prediction with lstm.
\newblock \emph{ICANN}.

\bibitem[{Graves(2013)}]{graves2013generating}
Alex Graves. 2013.
\newblock Generating sequences with recurrent neural networks.
\newblock \emph{arXiv preprint arXiv:1308.0850}.

\bibitem[{Hochreiter and Schmidhuber(1997)}]{hochreiter1997long}
Sepp Hochreiter and J{\"u}rgen Schmidhuber. 1997.
\newblock Long short-term memory.
\newblock \emph{Neural computation}, 9(8):1735--1780.

\bibitem[{Huang and Riloff(2011)}]{huang-riloff-2011-peeling}
Ruihong Huang and Ellen Riloff. 2011.
\newblock \href {https://www.aclweb.org/anthology/P11-1114} {Peeling back the
  layers: Detecting event role fillers in secondary contexts}.
\newblock In \emph{Proceedings of the 49th Annual Meeting of the Association
  for Computational Linguistics: Human Language Technologies}, pages
  1137--1147, Portland, Oregon, USA. Association for Computational Linguistics.

\bibitem[{Huang and Riloff(2012)}]{huang2012modeling}
Ruihong Huang and Ellen Riloff. 2012.
\newblock Modeling textual cohesion for event extraction.
\newblock In \emph{Twenty-Sixth AAAI Conference on Artificial Intelligence}.

\bibitem[{Ji and Grishman(2008)}]{ji-grishman-2008-refining}
Heng Ji and Ralph Grishman. 2008.
\newblock \href {https://www.aclweb.org/anthology/P08-1030} {Refining event
  extraction through cross-document inference}.
\newblock In \emph{Proceedings of ACL-08: HLT}, pages 254--262, Columbus, Ohio.
  Association for Computational Linguistics.

\bibitem[{Jia et~al.(2019)Jia, Wong, and Poon}]{jia-etal-2019-document}
Robin Jia, Cliff Wong, and Hoifung Poon. 2019.
\newblock \href {https://doi.org/10.18653/v1/N19-1370} {Document-level n-ary
  relation extraction with multiscale representation learning}.
\newblock In \emph{Proceedings of the 2019 Conference of the North {A}merican
  Chapter of the Association for Computational Linguistics: Human Language
  Technologies, Volume 1 (Long and Short Papers)}, pages 3693--3704,
  Minneapolis, Minnesota. Association for Computational Linguistics.

\bibitem[{Joshi et~al.(2019)Joshi, Levy, Weld, and Zettlemoyer}]{joshi2019bert}
Mandar Joshi, Omer Levy, Daniel~S Weld, and Luke Zettlemoyer. 2019.
\newblock Bert for coreference resolution: Baselines and analysis.
\newblock \emph{arXiv preprint arXiv:1908.09091}.

\bibitem[{Lafferty et~al.(2001)Lafferty, McCallum, and
  Pereira}]{lafferty2001conditional}
John Lafferty, Andrew McCallum, and Fernando~CN Pereira. 2001.
\newblock Conditional random fields: Probabilistic models for segmenting and
  labeling sequence data.
\newblock In \emph{ICML}.

\bibitem[{Lample et~al.(2016)Lample, Ballesteros, Subramanian, Kawakami, and
  Dyer}]{lample-etal-2016-neural}
Guillaume Lample, Miguel Ballesteros, Sandeep Subramanian, Kazuya Kawakami, and
  Chris Dyer. 2016.
\newblock \href {https://doi.org/10.18653/v1/N16-1030} {Neural architectures
  for named entity recognition}.
\newblock In \emph{Proceedings of the 2016 Conference of the North {A}merican
  Chapter of the Association for Computational Linguistics: Human Language
  Technologies}, pages 260--270, San Diego, California. Association for
  Computational Linguistics.

\bibitem[{Li et~al.(2013)Li, Ji, and Huang}]{li-etal-2013-joint}
Qi~Li, Heng Ji, and Liang Huang. 2013.
\newblock \href {https://www.aclweb.org/anthology/P13-1008} {Joint event
  extraction via structured prediction with global features}.
\newblock In \emph{Proceedings of the 51st Annual Meeting of the Association
  for Computational Linguistics (Volume 1: Long Papers)}, pages 73--82, Sofia,
  Bulgaria. Association for Computational Linguistics.

\bibitem[{Li et~al.(2015)Li, Nguyen, Cao, and
  Grishman}]{li-etal-2015-improving-event}
Xiang Li, Thien~Huu Nguyen, Kai Cao, and Ralph Grishman. 2015.
\newblock \href {https://doi.org/10.18653/v1/W15-4502} {Improving event
  detection with abstract meaning representation}.
\newblock In \emph{Proceedings of the First Workshop on Computing News
  Storylines}, pages 11--15, Beijing, China. Association for Computational
  Linguistics.

\bibitem[{Liao and Grishman(2010)}]{liao-grishman-2010-using}
Shasha Liao and Ralph Grishman. 2010.
\newblock \href {https://www.aclweb.org/anthology/P10-1081} {Using document
  level cross-event inference to improve event extraction}.
\newblock In \emph{Proceedings of the 48th Annual Meeting of the Association
  for Computational Linguistics}, pages 789--797, Uppsala, Sweden. Association
  for Computational Linguistics.

\bibitem[{Liu et~al.(2017)Liu, Chen, Liu, and Zhao}]{liu-etal-2017-exploiting}
Shulin Liu, Yubo Chen, Kang Liu, and Jun Zhao. 2017.
\newblock \href {https://doi.org/10.18653/v1/P17-1164} {Exploiting argument
  information to improve event detection via supervised attention mechanisms}.
\newblock In \emph{Proceedings of the 55th Annual Meeting of the Association
  for Computational Linguistics (Volume 1: Long Papers)}, pages 1789--1798,
  Vancouver, Canada. Association for Computational Linguistics.

\bibitem[{Liu et~al.(2019)Liu, Huang, and Zhang}]{liu-etal-2019-open}
Xiao Liu, Heyan Huang, and Yue Zhang. 2019.
\newblock \href {https://doi.org/10.18653/v1/P19-1276} {Open domain event
  extraction using neural latent variable models}.
\newblock In \emph{Proceedings of the 57th Annual Meeting of the Association
  for Computational Linguistics}, pages 2860--2871, Florence, Italy.
  Association for Computational Linguistics.

\bibitem[{Liu et~al.(2018)Liu, Luo, and Huang}]{liu-etal-2018-jointly}
Xiao Liu, Zhunchen Luo, and Heyan Huang. 2018.
\newblock \href {https://doi.org/10.18653/v1/D18-1156} {Jointly multiple events
  extraction via attention-based graph information aggregation}.
\newblock In \emph{Proceedings of the 2018 Conference on Empirical Methods in
  Natural Language Processing}, pages 1247--1256, Brussels, Belgium.
  Association for Computational Linguistics.

\bibitem[{Mintz et~al.(2009)Mintz, Bills, Snow, and
  Jurafsky}]{mintz-etal-2009-distant}
Mike Mintz, Steven Bills, Rion Snow, and Daniel Jurafsky. 2009.
\newblock \href {https://www.aclweb.org/anthology/P09-1113} {Distant
  supervision for relation extraction without labeled data}.
\newblock In \emph{Proceedings of the Joint Conference of the 47th Annual
  Meeting of the {ACL} and the 4th International Joint Conference on Natural
  Language Processing of the {AFNLP}}, pages 1003--1011, Suntec, Singapore.
  Association for Computational Linguistics.

\bibitem[{MUC-4(1992)}]{muc-1992-message}
MUC-4. 1992.
\newblock \href {https://www.aclweb.org/anthology/M92-1000} {Fourth message
  understanding conference ({MUC}-4)}.
\newblock In \emph{Proceedings of FOURTH MESSAGE UNDERSTANDING CONFERENCE
  ({MUC}-4)}, McLean, Virginia.

\bibitem[{Nguyen et~al.(2016)Nguyen, Cho, and
  Grishman}]{nguyen-etal-2016-joint}
Thien~Huu Nguyen, Kyunghyun Cho, and Ralph Grishman. 2016.
\newblock \href {https://doi.org/10.18653/v1/N16-1034} {Joint event extraction
  via recurrent neural networks}.
\newblock In \emph{Proceedings of the 2016 Conference of the North {A}merican
  Chapter of the Association for Computational Linguistics: Human Language
  Technologies}, pages 300--309, San Diego, California. Association for
  Computational Linguistics.

\bibitem[{Nguyen and Grishman(2015)}]{nguyen-grishman-2015-event}
Thien~Huu Nguyen and Ralph Grishman. 2015.
\newblock \href {https://doi.org/10.3115/v1/P15-2060} {Event detection and
  domain adaptation with convolutional neural networks}.
\newblock In \emph{Proceedings of the 53rd Annual Meeting of the Association
  for Computational Linguistics and the 7th International Joint Conference on
  Natural Language Processing (Volume 2: Short Papers)}, pages 365--371,
  Beijing, China. Association for Computational Linguistics.

\bibitem[{Patwardhan and Riloff(2009)}]{patwardhan-riloff-2009-unified}
Siddharth Patwardhan and Ellen Riloff. 2009.
\newblock \href {https://www.aclweb.org/anthology/D09-1016} {A unified model of
  phrasal and sentential evidence for information extraction}.
\newblock In \emph{Proceedings of the 2009 Conference on Empirical Methods in
  Natural Language Processing}, pages 151--160, Singapore. Association for
  Computational Linguistics.

\bibitem[{Peng et~al.(2017)Peng, Poon, Quirk, Toutanova, and
  Yih}]{peng-etal-2017-cross}
Nanyun Peng, Hoifung Poon, Chris Quirk, Kristina Toutanova, and Wen-tau Yih.
  2017.
\newblock \href {https://doi.org/10.1162/tacl_a_00049} {Cross-sentence n-ary
  relation extraction with graph {LSTM}s}.
\newblock \emph{Transactions of the Association for Computational Linguistics},
  5:101--115.

\bibitem[{Pennington et~al.(2014)Pennington, Socher, and
  Manning}]{pennington-etal-2014-glove}
Jeffrey Pennington, Richard Socher, and Christopher Manning. 2014.
\newblock \href {https://doi.org/10.3115/v1/D14-1162} {{G}love: Global vectors
  for word representation}.
\newblock In \emph{Proceedings of the 2014 Conference on Empirical Methods in
  Natural Language Processing ({EMNLP})}, pages 1532--1543, Doha, Qatar.
  Association for Computational Linguistics.

\bibitem[{Peters et~al.(2018)Peters, Neumann, Iyyer, Gardner, Clark, Lee, and
  Zettlemoyer}]{peters-etal-2018-deep}
Matthew Peters, Mark Neumann, Mohit Iyyer, Matt Gardner, Christopher Clark,
  Kenton Lee, and Luke Zettlemoyer. 2018.
\newblock \href {https://doi.org/10.18653/v1/N18-1202} {Deep contextualized
  word representations}.
\newblock In \emph{Proceedings of the 2018 Conference of the North {A}merican
  Chapter of the Association for Computational Linguistics: Human Language
  Technologies, Volume 1 (Long Papers)}, pages 2227--2237, New Orleans,
  Louisiana. Association for Computational Linguistics.

\bibitem[{Peters et~al.(2019)Peters, Ruder, and Smith}]{peters-etal-2019-tune}
Matthew~E. Peters, Sebastian Ruder, and Noah~A. Smith. 2019.
\newblock \href {https://doi.org/10.18653/v1/W19-4302} {To tune or not to tune?
  adapting pretrained representations to diverse tasks}.
\newblock In \emph{Proceedings of the 4th Workshop on Representation Learning
  for NLP (RepL4NLP-2019)}, pages 7--14, Florence, Italy. Association for
  Computational Linguistics.

\bibitem[{Sundheim(1992)}]{sundheim-1992-overview}
Beth~M. Sundheim. 1992.
\newblock \href {https://www.aclweb.org/anthology/M92-1001} {Overview of the
  fourth message understanding evaluation and conference}.
\newblock In \emph{FOURTH MESSAGE UNDERSTANDING CONFERENCE ({MUC}-4),
  Proceedings of a Conference Held in McLean, Virginia, June 16-18, 1992}.

\bibitem[{Trinh et~al.(2018)Trinh, Dai, Luong, and Le}]{pmlr-v80-trinh18a}
Trieu Trinh, Andrew Dai, Thang Luong, and Quoc Le. 2018.
\newblock \href {http://proceedings.mlr.press/v80/trinh18a.html} {Learning
  longer-term dependencies in {RNN}s with auxiliary losses}.
\newblock In \emph{Proceedings of the 35th International Conference on Machine
  Learning}, volume~80 of \emph{Proceedings of Machine Learning Research},
  pages 4965--4974, Stockholmsmässan, Stockholm Sweden. PMLR.

\bibitem[{Vaswani et~al.(2017)Vaswani, Shazeer, Parmar, Uszkoreit, Jones,
  Gomez, Kaiser, and Polosukhin}]{vaswani2017attention}
Ashish Vaswani, Noam Shazeer, Niki Parmar, Jakob Uszkoreit, Llion Jones,
  Aidan~N Gomez, {\L}ukasz Kaiser, and Illia Polosukhin. 2017.
\newblock Attention is all you need.
\newblock In \emph{Advances in neural information processing systems}, pages
  5998--6008.

\bibitem[{Wadden et~al.(2019)Wadden, Wennberg, Luan, and
  Hajishirzi}]{wadden-etal-2019-entity}
David Wadden, Ulme Wennberg, Yi~Luan, and Hannaneh Hajishirzi. 2019.
\newblock \href {https://doi.org/10.18653/v1/D19-1585} {Entity, relation, and
  event extraction with contextualized span representations}.
\newblock In \emph{Proceedings of the 2019 Conference on Empirical Methods in
  Natural Language Processing and the 9th International Joint Conference on
  Natural Language Processing (EMNLP-IJCNLP)}, pages 5788--5793, Hong Kong,
  China. Association for Computational Linguistics.

\bibitem[{Yang and Mitchell(2016)}]{yang-mitchell-2016-joint}
Bishan Yang and Tom~M. Mitchell. 2016.
\newblock \href {https://doi.org/10.18653/v1/N16-1033} {Joint extraction of
  events and entities within a document context}.
\newblock In \emph{Proceedings of the 2016 Conference of the North {A}merican
  Chapter of the Association for Computational Linguistics: Human Language
  Technologies}, pages 289--299, San Diego, California. Association for
  Computational Linguistics.

\bibitem[{Zhao et~al.(2018)Zhao, Jin, Wang, and
  Cheng}]{zhao-etal-2018-document}
Yue Zhao, Xiaolong Jin, Yuanzhuo Wang, and Xueqi Cheng. 2018.
\newblock \href {https://doi.org/10.18653/v1/P18-2066} {Document embedding
  enhanced event detection with hierarchical and supervised attention}.
\newblock In \emph{Proceedings of the 56th Annual Meeting of the Association
  for Computational Linguistics (Volume 2: Short Papers)}, pages 414--419,
  Melbourne, Australia. Association for Computational Linguistics.

\end{thebibliography}
\bibliographystyle{acl_natbib}

\end{document}